\documentclass[10pt,twocolumn,letterpaper]{article}
\usepackage[pagenumbers]{iccv} 

\definecolor{iccvblue}{rgb}{0.21,0.49,0.74}
\usepackage[pagebackref,breaklinks,colorlinks,allcolors=iccvblue]{hyperref}
\usepackage{makecell}
\usepackage{algorithm}
\usepackage{algorithmic}
\usepackage{multirow}
\usepackage{arydshln}
\usepackage[T1]{fontenc} 
\def\confName{ICCV}
\def\confYear{2025}

\title{\textcolor[rgb]{0.7529,0.2117,0.1921}{i}\textcolor[rgb]{0.8196,0.5215,0.3294}{Manip}: Skill-\textcolor[rgb]{0.7529,0.2117,0.1921}{I}ncremental Learning for Robotic \textcolor[rgb]{0.8196,0.5215,0.3294}{Manip}ulation}

\author{
Zexin Zheng\textsuperscript{1}, Jia-Feng Cai\textsuperscript{1}, Xiao-Ming Wu\textsuperscript{1}, Yi-Lin Wei \textsuperscript{1}, Yu-Ming Tang\textsuperscript{1}, Wei-Shi Zheng\textsuperscript{1,2,3} \\
\textsuperscript{1}School of Computer Science and Engineering, Sun Yat-sen University, China, \textsuperscript{2}Pengcheng Lab, China \\
\textsuperscript{3}Key Laboratory of Machine Intelligence and Advanced Computing, Ministry of Education, China\\
{\tt\small \{zhengzx25, caijf23, wuxm65, weiylin5, tangym9\}@mail2.sysu.edu.cn}
,\tt\small wszheng@ieee.org}

\begin{document}
\maketitle
\vspace{-1mm}
\begin{abstract}
The development of a generalist agent with adaptive multiple manipulation skills has been a long-standing goal in the robotics community.
In this paper, we explore a crucial task, \textbf{skill-incremental learning}, in robotic manipulation, which is to endow the robots with the ability to learn new manipulation skills based on the previous learned knowledge without re-training. First, we build a skill-incremental environment based on the RLBench benchmark, and explore how traditional incremental methods perform in this setting. We find that they suffer from severe catastrophic forgetting due to the previous methods on classification overlooking the characteristics of temporality and action complexity in robotic manipulation tasks. Towards this end, we propose an \textbf{i}ncremental \textbf{Manip}ulation framework, termed \textbf{iManip}, to mitigate the above issues. We firstly design a temporal replay strategy to maintain the integrity of old skills when learning new skill. Moreover, we propose the extendable PerceiverIO, consisting of an action prompt with extendable weight to adapt to new action primitives in new skill. Extensive experiments show that our framework performs well in Skill-Incremental Learning. Codes of the skill-incremental environment with our framework will be open-source.  
\end{abstract}

\vspace{-2mm}
\section{Introduction}
\label{sec:intro}

\vspace{-2mm}
\begin{figure}[t]
  \centering
   \includegraphics[width=1\linewidth]{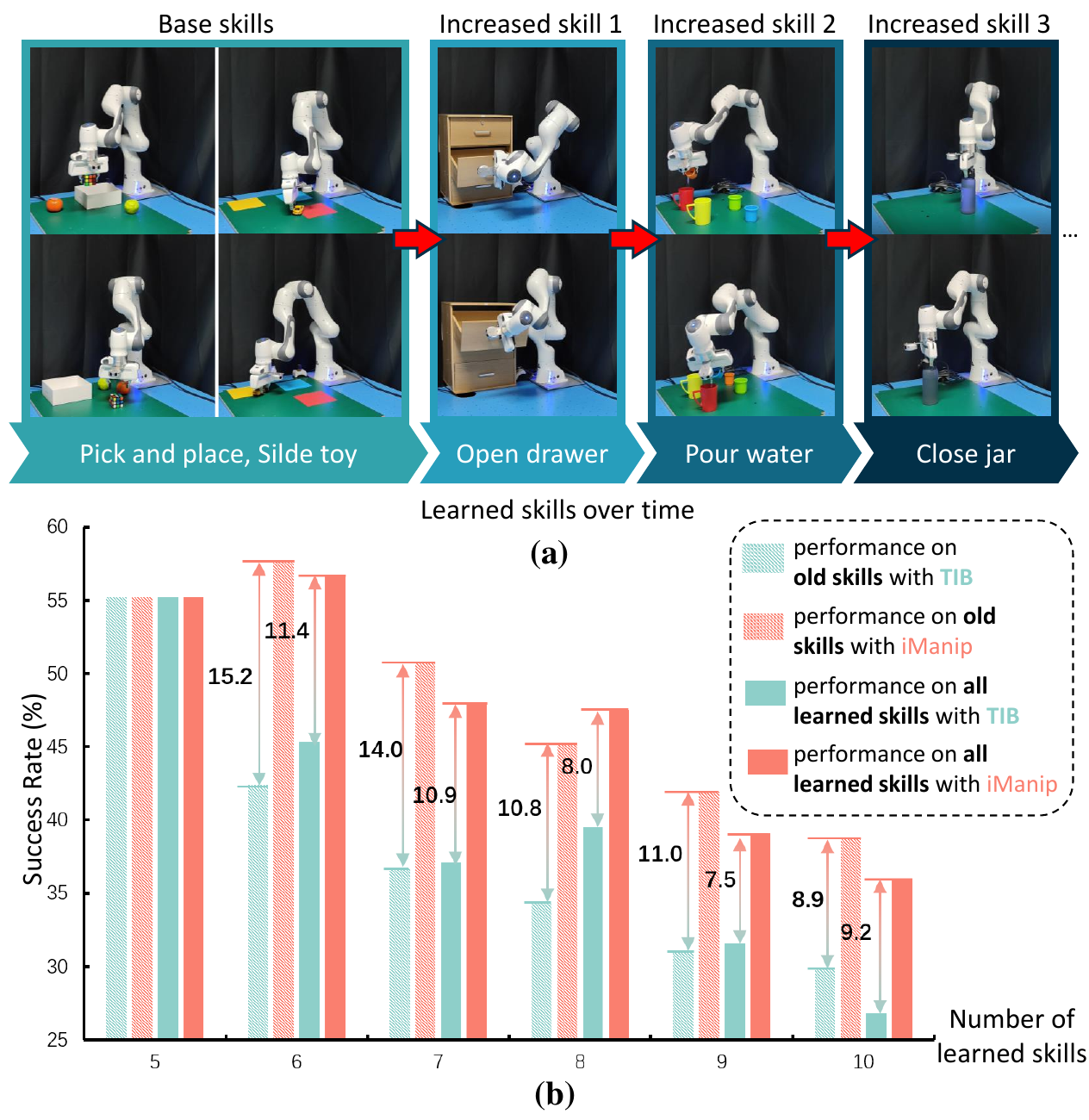}

   \caption{(a) An overview of our skill-incremental learning for robotic manipulation that requires the agent to learn skill sequences over time.
   (b) A comparison of model performance between the traditional incremental baseline (TIB) and our iManip.
   }\vspace{-3mm}
   \label{fig:fig1}
\end{figure}

Imagine that we are in a household setting with a robot assistant that already has basic functions like folding clothes and fetching items. Now, as the owners, we want it to learn new skills. For instance, today we’ve purchased a dishwasher, and we’d like the robot to learn how to load dishes into it. It could quickly acquire these new skills by observing and mimicking our demonstrations. This is an interesting and challenging requirement for robotics manipulation, which needs the robot to learn new skills based on previously learned knowledge without retraining. However, in the area of robotic manipulation, previous research mainly focuses on how to acquire better manipulation performance \cite{jaegle2021perceiver, tsurumine2022goal, ryu2024diffusion, ma2024hierarchical, goyal2023rvt, cai2024real} or how to transfer the knowledge from the pretrained large language or vision models \cite{black2024pi_0, zawalski2024robotic, kim2024openvla, cheang2024gr, huang2024rekep} to learn robotics manipulations, rarely works explore how to incrementally learn new skills. The LIBERO \cite{liu2024libero} is a preliminary benchmark to learn lifelong robotics control, but it explores the incremental abilities in new objects or new spatial positions, still limited in the characteristics of the same skills, where different tasks share the same skill.

In this paper, we thoroughly explore this crucial task, \textbf{skill-incremental learning}, in robotic manipulation, which is to leverage the previous learned knowledge to enable the robots to learn new manipulation skills without training from scratch. To begin with, we construct a skill-incremental environment based on the RLBench benchmark. It requires the agent to continuously acquire a sequence of 10 challenging language-conditioned manipulation skills. Each skill consists of at least two variations encompassing several types, such as variations in shape and color, totaling 166 variations. Then, we explore how the traditional incremental learning methods work on this skill-incremental environment. As seen in \Cref{fig:fig1}, we discover that after learning new manipulation skills, the agent's performance on prior skills significantly deteriorates. Thus, we can conclude that traditional incremental learning methods still suffer from catastrophic forgetting in this new setting, which, we regard, is due to their neglect of the temporality and action complexity in robotic manipulation tasks. The temporal complexity arises from the dynamic changes in the environment and in the robot states over time, resulting in each action having an impact on subsequent actions. And the complexity of actions requires the agent to learn new action primitives for action planning in novel environments and interactions, which is representative in 3D dimensions with rotation and shift and highly complex.

To mitigate the above problems, we propose a new skill \textbf{i}ncremental \textbf{Manip}ulation framework, termed \textbf{iManip}, for this new setting. The key idea of our framework is to modify the traditional incremental learning methods to fully consider the characteristics of temporality and action complexity in robotic manipulation tasks. Two designs make our framework nontrivial. First, to address temporal complexity, we design the temporal replay strategy to maintain the integrity of the temporal data and propose to replay a fixed number of keyframe samples at different time points for each manipulation skill, using the farthest-distance entropy sampling strategy. Second, to address the complexity of actions, we propose the Extendable PerceiverIO, consisting of an action prompt with extendable weight to adapt to new action primitives. When learning new skills, we freeze the learned parameters of PerceiverIO while learning a new small set of skill-specific action prompts and weight matrices for new action primitives learning.

Extensive experiments show that our iManip framework maintains several excellent capabilities: 
(1) \textbf{Effectiveness}: it performs well in the skill-incremental learning setting, outperforming the traditional incremental baseline with a \textbf{9.4} increase in average. 
(2) \textbf{Robustness}: it demonstrates robustly performance in several different incremental settings.
(3) \textbf{Lightweight}: it only needs lightweight finetuning of the policy decoder with fewer training steps, comparable with full weights finetuning.
(4) \textbf{Extendability}: it also has extraordinary performance in real-world experiments.
Codes of the skill-incremental environment with our framework will be open-source.

\section{Related Work}
\noindent\textbf{Robotic Manipulation.} 
Learning robot manipulation conditioned on both vision and language has gained increasing attention~\cite{shao2021concept2robot,stepputtis2020language,chen2023polarnet,goyal2023rvt,guhur2023instruction,gervet2023act3d,shridhar2022cliport,xian2023chaineddiffuser}, with robot imitation learning using scripted trajectories~\cite{james2020rlbench, mees2022calvin} or tele-operation data~\cite{o2023open, cheng2024open, wu2024gello} gradually becoming a mainstream approach. 
Previous work~\cite{jang2022bc,brohan2022rt,chi2023diffusion} has focused on using 2D images to predict actions, while more recent studies have leveraged the rich spatial information of 3D point clouds for motion planning. 
For instance,  
PerAct~\cite{shridhar2023perceiver} feeds voxel tokens into a PerceiverIO~\cite{jaegle2021perceiver}-based transformer policy, achieving impressive results across various tasks.
GNFactor~\cite{ze2023gnfactor} optimizes a generalizable neural field for semantic extraction, while ManiGaussian~\cite{lu2024manigaussian} introduces a dynamic Gaussian Splatting~\cite{kerbl20233d} framework for semantic propagation. 
3DDA~\cite{ke20243d} proposes a 3D denoising transformer to predict noise in noised 3D robot pose trajectories.
However, these methods suffer from catastrophic forgetting in skill-incremental learning and we propose our iManip framework to continuously learn new skills and mitigate forgetting of learned knowledge.

\noindent\textbf{Conventional Lifelong Learning Approaches.}
One of the most commonly used methods is the rehearsal-based method~\cite{chaudhry2019tiny,rebuffi2017icarl,chaudhry2018riemannian,hou2019learning,DDE,afc,GCR,CSCCT}.
iCaRL~\cite{rebuffi2017icarl} proposes a herding-based step for prioritized exemplar selection to store old exemplars.
RWalk~\cite{chaudhry2018riemannian} proposes a hard-exemplar sampling strategy for replay.
Distillation-based methods~\cite{li2017learning,hou2019learning,douillard2020podnet,castro2018end,lee2019overcoming,smith2021always,zhou2021co} propose distilling old knowledge from the old network to the current network or maintaining the old feature space during new tasks.
ABD~\cite{smith2021always} proposes distilling synthetic data for incremental learning and EEIL~\cite{castro2018end} proposes to distill the knowledge from the classification layers of the old classes.
Moreover, there are dynamic-architecture-based methods~\cite{ostapenko2019learning,li2019learn,pham2021dualnet,yan2021dynamically,liu2021adaptive,aljundi2017expert} that dynamically adjust the model’s representation ability to fit the evolving data stream.
In this work, we use the traditional rehearsal-based method~\cite{rebuffi2017icarl} and the distillation-based methods~\cite{castro2018end} for robotic skill-incremental learning.
We find that they also suffer from catastrophic forgetting due to overlooking the temporal and action complexities of robotic manipulation.

\begin{figure*}[t]
  \centering
   \includegraphics[width=1\linewidth]{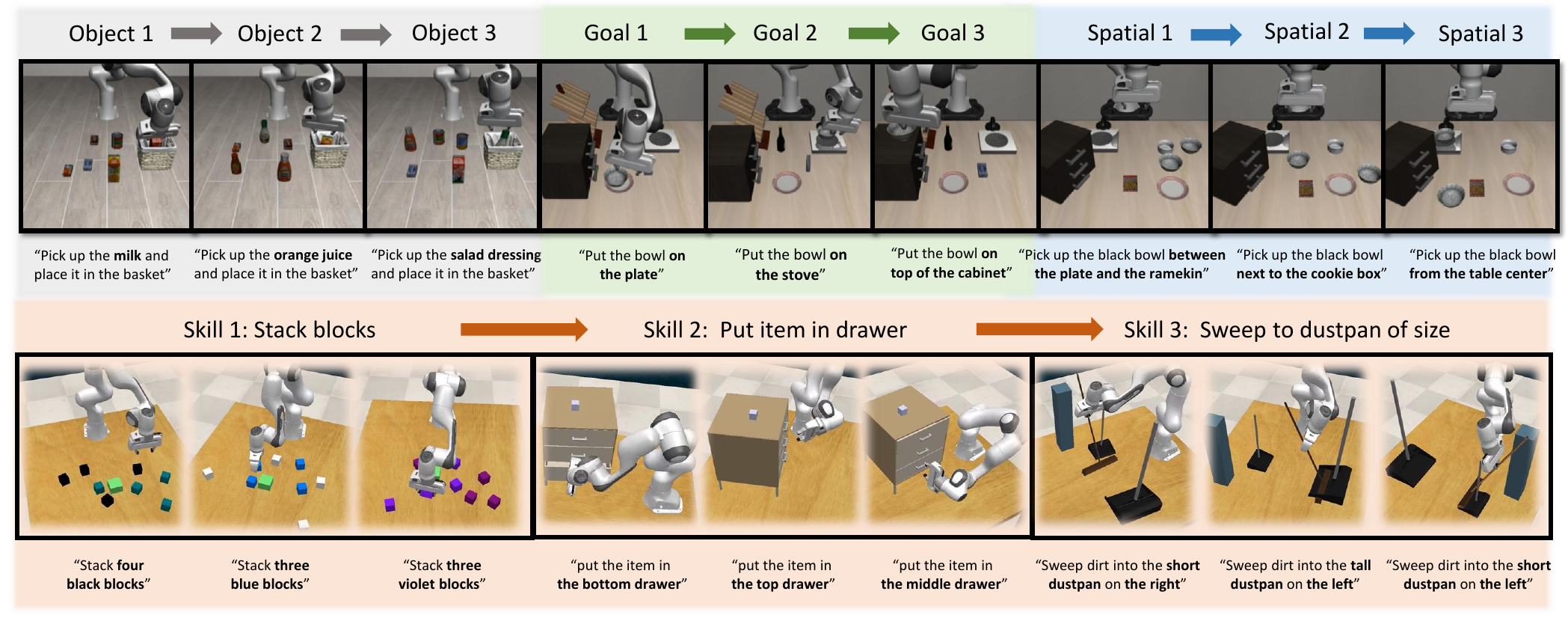}
\caption{Overview of robotic incremental learning. Previous works focus on incremental abilities in new objects, goals, or spatial positions, where different tasks may share the same skill. The iManip focuses on skill-incremental learning which better captures the true adaptability and flexibility required for real-world robotic learning.}\vspace{-4mm}
\label{fig:fig2}
\end{figure*}

\noindent\textbf{A Preliminary Lifelong Robot Learning Benchmark LIBERO.} Previous works \cite{kim2024online, khetarpal2018environments, wolczyk2021continual, mendez2018lifelong, gao2021cril, meng2025preserving} explore different strategies for incremental robotic learning based on different testing environments. Recently, to promote community development, LIBERO~\cite{liu2024libero} proposed a benchmark, which explores incremental abilities with new objects, goals, or spatial positions, as shown in \Cref{fig:fig2}. The limitation of LIBERO lies in the fact that most tasks are constrained by similar skill characteristics. For example, it regards ``Put the bowl on the plate" and ``Put the bowl on stove" as different tasks. In this paper, we explore a more realistic and challenging task, skill-incremental learning, where the agent learns a sequence of skills over time, each involving multiple poses and object variations in placement, color, shape, size, and category.

\section{Skill-incremental Learning for Robotic Manipulation}

\subsection{Challenges}
While robotic manipulation has received increasing attention, few previous studies explore how to incrementally learn new skills.
In this paper, we focus on skill-incremental learning for robotic manipulation, a challenging setting that requires agents to continuously acquire new skills without training from scratch.

In this new setting, we find that applying traditional incremental learning methods~\cite{castro2018end,rebuffi2017icarl,chaudhry2018riemannian,lu2024manigaussian,shridhar2023perceiver} still suffers from catastrophic forgetting of previously learned skills, as seen in ~\Cref{fig:fig1} (b).
There are two key challenges when applying previous methods of visual classification to robotic skill-incremental learning:
1) Previous methods overlook the temporal complexity inherent in robotic manipulation tasks, where dynamic changes in the environment and the robot states over time cause actions to impact subsequent ones.
For example, classical replay algorithms focus on sampling the most representative samples per class, directly storing representative samples from demonstrations may result in temporal imbalance of the trajectory, leading to instability during task execution.
2) Previous methods focus primarily on general visual features while neglecting the actions complexity in robotic manipulation. 
Robotic manipulation involves action planning through interactions with the physical environment, such as visual and language input.
When a new manipulation skill arises, the agent learns new visual-language interactions and quickly acquires new action primitives based on prior knowledge.
\begin{figure*}[t]
  \centering
   \includegraphics[width=1\linewidth]{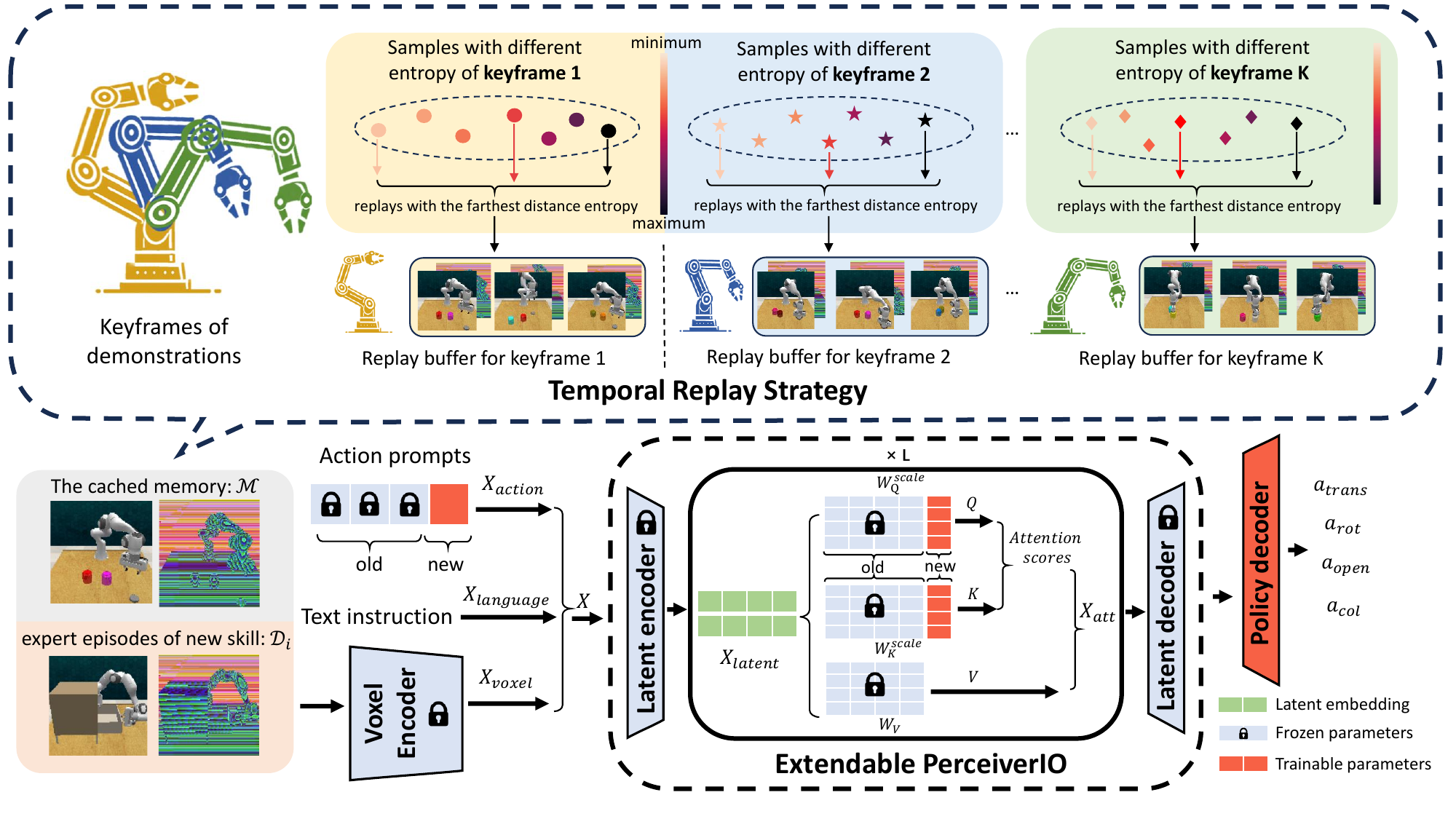}

   \caption{
   The overall framework of iManip, which primarily consists of a temporal replay strategy to store the samples with the farthest distance entropy for each keyframe of old demonstrations and an extendable PerceiverIO consisting of action prompts with extendable weights to adapt to new action primitives.
   }
   \label{fig:fig3}
\end{figure*}

\subsection{Overall Pipeline}
To tackle the two challenges, we propose the temporal replay strategy and the extendable PerceiverIO architecture in our iManip framework. Specifically, as shown in \Cref{fig:fig3}, we present the overall framework for robotic skill-incremental learning, which can sequentially learn robotic manipulation skills while mitigating catastrophic forgetting of learned skills. 
Specifically, the agent learns a sequence of manipulation skills with a stream of training data denoted as $\mathcal{D}=\{\mathcal{D}_{i}\}_{i=1}^{T}$, where $\mathcal{D}_i=\{(o^{(1)}_i,a^{(1)}_i),(o^{(2)}_i,a^{(2)}_i),\ldots\}$ represents the demonstration trajectories of skill $i$.
The visual input $o^{(t)}_i = (I^{(t)}_i, D^{(t)}_i, P^{(t)}_i)$ consists of the $t$-th single-view images $I^{(t)}_i$, depth images $D^{(t)}_i$, and proprioception matrix $P^{(t)}_i \in \mathbb{R}^4$ that includes the openness, end-effector position, and the current timestep.
After learning skill $i$, a compact memory $\mathcal{M}$ stores a fixed number of demonstration replays for skills up to $i-1$.
Following ~\cite{lu2024manigaussian,shridhar2023perceiver,ze2023gnfactor},
the agent combines the visual input $o^{(t)}_i$ and language instructions $l_i$ to generates the optimal action $a^{(t)}_i=(a^{(t)}_{i,{\text{trans}}}, a^{(t)}_{i,{\text{rot}}},
a^{(t)}_{i,{\text{open}}},
a^{(t)}_{i,{\text{col}}})$, which respectively demonstrates the target translation in voxel $a^{(t)}_{i,{\text{trans}}} \in \mathbb{R}^{100^3}$, rotation $a^{(t)}_{i,{\text{rot}}} \in \mathbb{R}^{(360/5) \times 3}$, openness $a^{(t)}_{i,{\text{open}}} \in [0, 1]$ and collision avoidance $a^{(t)}_{i,{\text{col}}} \in [0, 1]$.

Our framework consists of a voxel encoder for learning 3D scene features, a latent transformer (the extendable PerceiverIO), and a policy decoder to predict optimal robot actions.
Specifically, our approach employs a temporal replay strategy, maximizing the information entropy of replay demos to effectively address the first challenges caused by classic replay methods. 
Additionally, we introduce the extendable PerceiverIO, consisting of action prompts with extendable weights to
adapt to new action primitives, while preserving knowledge of previous skills to prevent catastrophic forgetting.

\subsection{The iManip Framework}
\noindent\textbf{Temporal Replay Strategy.}
Rehearsal-based methods~\cite{rebuffi2017icarl,hou2019learning,chaudhry2018riemannian} are classic algorithms in incremental vision classification tasks while overlooking the temporal complexity inherent in robotic manipulation. 
Apart from random sampling, popular methods such as herding sampling~\cite{rebuffi2017icarl} and hard-exemplar sampling~\cite{chaudhry2018riemannian,bang2021rainbow} effectively select the most representative samples from each class.
However, robotic manipulation data consists of temporal samples from entire expert episodes. 
Directly storing representative samples can lead to temporal imbalance of episodes, resulting in instability during task execution.

Therefore, we propose a temporal replay strategy to balance the sampling of different keyframes of episodes for each skill.
Keyframes~\cite{shridhar2023perceiver} are the samples from episodes when the end-effector changes state (e.g., gripper closing) or when its velocity approaches zero, representing critical temporal landmarks within the trajectory.

Furthermore, to sample a greater variety of variants, we propose the farthest-distance entropy sampling to store an equal number of each keyframe.
It requires the buffer to contain the replays that exhibit the largest entropy divergence as
\begin{equation}
  \begin{gathered} S= \mathop{\arg\max}\limits_{S \in E,|S|=K}\sum_{i\in S}\sum_{j\in S}A[i][j],
 \end{gathered}
  \label{eq:3.0}
\end{equation}
\\
where $S$ is the sampled set with size $K$, $E$ is the demo set of a specific keyframe with size $N$, and $A$ is the distance array of the demos' action prediction entropy $\mathcal{L}_{\text{act}}$.
Specifically, we propose to store the demo $j$ to the replay buffer that has the farthest distance of sampled demos in $S$ as 
\begin{equation}
  \begin{gathered} j = \mathop{\arg\max}\limits_{j \in E}\sum_{k\in S}A[j][k].
 \end{gathered}
  \label{eq:3.0.1}
\end{equation}
This ensures the storage of temporally balanced, information-rich samples from previous episodes, helping to mitigate catastrophic forgetting of learned skills.

Based on the above analysis, the pseudocode for the temporal replay strategy is shown in \Cref{alg:fps}. The algorithm can get the optimal solution for the objective \ref{eq:3.0}, with time complexity of $O(N^2)$, which is the size of a specific keyframe and not relevant to the size of the previous data, suitable for incremental skill learning.

\begin{algorithm}[H]
\caption{Farthest-distance Entropy Sampling}
\label{alg:fps}
\begin{algorithmic}[1]
\REQUIRE Entropy set corresponding to the keyframe samples $E = \{e_1, e_2, \dots, e_N\}$, sampling size $K$
\ENSURE Sampled set $S = \{s_1, s_2, \dots, s_K\}$

\STATE Calculate the distance array $A$, where $A[i][j] = \text{distance}(e_i, e_j)$
\STATE Select the sample $i$ and add to $S$, where
$i = \mathop{\arg\max}\limits_{1 \leq i \leq N} \sum_{j=1}^{N} A[i][j]$

\FOR{$k = 2$ to $K$}
    \STATE Find the value in $E$ that has the largest difference with the values in $S$, \\ $j = \mathop{\arg\max}\limits_{j \in E}\sum_{k\in S}A[j][k]$
    \STATE Add sample $j$ to $S$ as the $k$-th sample
\ENDFOR

\RETURN Sampled set $S$
\end{algorithmic}
\end{algorithm}

\noindent\textbf{Extendable PerceiverIO.}
Unlike traditional vision classification tasks, robotic manipulation requires the integration of multiple modalities to enable complex decision-making and long-term action planning through interactions with the physical environment.
Classic methods~\cite{rebuffi2017icarl,castro2018end,chaudhry2018riemannian,hou2019learning} for lifelong classification overlook the complexities of action in robotic manipulation tasks that require the agent to learn various action primitives for different skills. 
In our iManip framework, we propose an extendable PerceiverIO, which learns skill-specific action prompts with extendable weights to adapt to new action primitives.

Specifically, as illustrated in \Cref{fig:fig3}, the input to the extendable PerceiverIO consists of multimodal patches, $X=[X_{\text{voxel}},X_{\text{language}},X_{\text{action}}]$, where $X_{\text{voxel}}$  and $X_{\text{language}}$ represent the input sequences of voxel and language encodings, respectively, and $X_{\text{action}}=[X_{\text{action}}^{\text{old}},X_{\text{action}}^{\text{new}}]$ is the skill-specific action prompt that concatenates both the old and new action prompts. 
The notation $[\cdot,\cdot]$ denotes the concatenation operation along the token dimension.
Subsequently, $X$ undergoes a cross-attention computation between the input and a much smaller set of latent vectors through the latent encoder, producing $X_{\text{latent}}$.
Then $X_{\text{latent}}$ is encoded with weight-extendable self-attention layers as
\begin{equation}
  \begin{gathered} Q=X_{\text{latent}}\cdot W^{\text{scale}}_Q,\quad K=X_{\text{latent}}\cdot W^{\text{scale}}_K, \\
  \quad V=X_{\text{latent}}\cdot W_V,
 \end{gathered}
  \label{eq:3.1}
\end{equation}
\begin{equation}
  \begin{gathered} X_{\text{att}}=\mathrm{softmax}[\frac{Q\cdot K^\top}{\sqrt{d}}]\cdot V,
 \end{gathered}
  \label{eq:3.2}
\end{equation}
\noindent where $X_{\text{latent}},X_{\text{att}}\in\mathbb{R}^{T\times d}$ are respectively a set of $T$ input and output tokens with channel dimension $d$, $W^{\text{scale}}_Q,W^{\text{scale}}_K\in\mathbb{R}^{d\times d'}$,$W_V\in\mathbb{R}^{d\times d}$ are learnable weight matrices. 
Notably, $W^{\text{scale}}_Q,W^{\text{scale}}_K$ are extendable by appending newly weight matrices $W^{\text{new}}_Q,W^{\text{new}}_K\in\mathbb{R}^{d\times d_{\text{new}}}$ as
\begin{equation}
  \begin{gathered} W^{\text{scale}}_Q=
\begin{bmatrix}
W^{\text{old}}_Q,W^{\text{new}}_Q
\end{bmatrix},\quad W^{\text{scale}}_K=
\begin{bmatrix}
W^{\text{old}}_K,W^{\text{new}}_K
\end{bmatrix},
 \end{gathered}
  \label{eq:3.3}
\end{equation}
\noindent where $W^{\text{old}}_Q,W^{\text{old}}_K\in\mathbb{R}^{d\times d_{\text{old}}}$ are the old weight matrices and the expanded dimension $d'=d_{\text{old}}+d_{\text{new}}$. 
Finally, these encoded latents are cross-attended with the input once again through the latent decoder to ensure alignment with the input size.
In our iManip framework, we freeze the old PerceiverIO while learning the action prompts $X_{\text{action}}^{\text{new}}$ and a small set of newly weight matrices $W^{\text{new}}_Q,W^{\text{new}}_K$ of new skills.
This enables the agent to quickly adapt to new action primitives while preventing the forgetting of previous skills.

\noindent\textbf{Knowledge distillation between the old and new agents.}
To better preserve the knowledge of previous skills while learning new ones, we employ knowledge distillation~\cite{castro2018end,wu2019large}, where the output probability distribution of the old model is used to train the new model. 
This enables the transfer of knowledge from the old to the new agent, as defined by the following objective:
\begin{equation}
  \begin{aligned} 
  \mathcal{L}_{\text{dis}} =\mathcal{L}_2(\mathcal{Q}_{\text{trans}}^{\text{old}},\mathcal{Q}_{\text{trans}}^{\text{new}})+\mathcal{L}_2(\mathcal{Q}_{\text{rot}}^{\text{old}},\mathcal{Q}_{\text{rot}}^{\text{new}})+\\
  |\mathcal{Q}_{\text{open}}^{\text{old}}-\mathcal{Q}_{\text{open}}^{\text{new}}|+|\mathcal{Q}_{\text{collide}}^{\text{old}}-\mathcal{Q}_{\text{collide}}^{\text{new}}|,
 \end{aligned}
  \label{eq:3.10}
\end{equation}
\noindent where $\mathcal{L}_2$ is the MSE loss,  $[\mathcal{Q}_{\text{trans}}^{\text{old}},\mathcal{Q}_{\text{rot}}^{\text{old}},\mathcal{Q}_{\text{open}}^{\text{old}},\mathcal{Q}_{\text{collide}}^{\text{old}}]$ and $[\mathcal{Q}_{\text{trans}}^{\text{new}},\mathcal{Q}_{\text{rot}}^{\text{new}},\mathcal{Q}_{\text{open}}^{\text{new}},\mathcal{Q}_{\text{collide}}^{\text{new}}] $ denote the probabilities of the ground truth actions in expert demonstrations for translation, rotation, gripper openness, and collision avoidance for the old and new robots, respectively.
\begin{table*}[ht]
\centering
\begin{tabular}{c|ccccccccccccc}
\Xhline{1pt}
\multirow{2}{*}{Methods} & \multirow{2}{*}{Base} & \multicolumn{2}{c}{Step   1} & \multicolumn{2}{c}{Step   2} & \multicolumn{2}{c}{Step   3} & \multicolumn{2}{c}{Step   4} & \multicolumn{2}{c}{Step   5} & \multicolumn{2}{c}{Average}   \\ & & Old & All  & Old & All  & Old & All & Old & All & Old & All& Old & \multicolumn{1}{c}{All} \\ \hline
\rowcolor{gray!10} 
\multicolumn{14}{l}{\textit{multi-task methods}}  \\ \hline
PerAct~\cite{shridhar2023perceiver} & 44.0&  4.0  & 7.3 & 2.7 & 5.1 &  1.1  &  9.0  & 2.5  & 6.7  &  1.3  &  1.6  &  2.3  & 5.9 \\
ManiGaussian~\cite{lu2024manigaussian} & 55.2&  12.0  & 20.7 & 6.7 & 12.0 &  5.7  &  15.5  & 3.0  & 9.3  &  5.3  &  5.2  &  6.5  &  12.5 \\ \hline
\rowcolor{gray!10} 
\multicolumn{14}{l}{\textit{skill-incremental methods}}  \\ \hline
P-TIB~\cite{shridhar2023perceiver,rebuffi2017icarl,castro2018end} &     44.0   &  33.6 &  34.7  & 26.0 &25.1 & 22.3 & 26.0 & 17.0 & 16.4 &  11.6  & 10.4 &  22.1 &  22.5  \\
M-TIB~\cite{lu2024manigaussian,rebuffi2017icarl,castro2018end} &     55.2   &  42.4 &  45.3  &  36.7 & 37.1 & 34.3 & 39.5 & 31.0 &     31.6 &  29.8&  26.8&  34.8 &  36.1  \\
\textbf{Ours (iManip)}  & \textbf{56.0} &   \textbf{57.6} &  \textbf{56.7} &\textbf{50.7}   & \textbf{48.0} & \textbf{45.1} & \textbf{47.5}&\textbf{42.0}  &  \textbf{39.1}  &  \textbf{38.7}  &  \textbf{36.0}  &  \textbf{46.8}  &  \textbf{45.5}  \\ \Xhline{1pt}
\end{tabular}
\caption{Performance comparison of different methods of B5-5N1 in Rlbench.
We show the average success rate of old and all learned skills, and the average performance of all new steps.
The traditional incremental methods~\cite{rebuffi2017icarl,castro2018end} on baseline~\cite{lu2024manigaussian,shridhar2023perceiver} is termed TIB.} \vspace{-2mm}
\label{tab:maintable}
\end{table*}

\subsection{Learning Objectives}
Our approach is to address the problem of skill-incremental learning for robotic manipulation from multiple aspects.
First, for each manipulation skill, there is an action loss to facilitate robot imitation learning.
Following~\cite{shridhar2023perceiver,ze2023gnfactor,lu2024manigaussian}, we employ cross-entropy loss (CE) to ensure accurate action prediction:
\vspace{-4mm}
\begin{equation}
  \begin{aligned}\\\mathcal{L}_{\text {act}}=-\mathbb{E}_{Y_{\text {trans }}}\left[\log \mathcal{V}_{\text {trans }}\right]-\mathbb{E}_{Y_{\text {rot }}} \left[\log \mathcal{V}_{\text {rot }}\right] \\ -\mathbb{E}_{Y_{\text {open }}}\left[\log \mathcal{V}_{\text {open }}\right]-\mathbb{E}_{Y_{\text {collide }}}\left[\log \mathcal{V}_{\text {collide }}\right],\\\end{aligned}
  \label{eq:3.10}
\end{equation}
\noindent where $\mathcal{V}_i=\operatorname{softmax}(\mathcal{Q}_i)$ for $\mathcal{Q}_i\in[\mathcal{Q}_{\mathrm{trans}},    \mathcal{Q}_{\mathrm{open}},\mathcal{Q}_{\mathrm{rot}},$
$\mathcal{Q}_{\mathrm{collide}}]$ and $Y_i\in[Y_{\mathrm{trans}},Y_{\mathrm{rot}},Y_{\mathrm{open}},Y_{\mathrm{collide}}]$ is the ground truth one-hot encoding.

Furthermore, when learning new skills, we propose the temporal replay strategy to preserve a fixed number of representative samples from old demonstrations. 
The cached memory $\mathcal{M}$ will be used in conjunction with the new skill demos 
$\mathcal{D}_{\text{new}}$ for learning new skills.
Additionally, our expandable PerceiverIO will dynamically expand new learnable weights for the new skill.
We find that training only the skill-specific action prompts $ X_{\text{action}}^{\text{new}}$ with newly appended weights $W^{\text{new}}_Q,W^{\text{new}}_K$ and the policy decoder effectively prevents catastrophic forgetting, more analysis can be seen in the 3rd experiments in \Cref{sec4.2}.
Finally, we employ knowledge distillation loss $\mathcal{L}_{\text{dis}}$ to help the agent retain the knowledge of previous skills.
Overall, in skill-incremental learning, our training loss is formulated as follows:
\begin{equation}
  \begin{aligned} 
  \mathcal{L}_{\text{total}} =\mathcal{L}_{\text{act}} + \lambda_{\text{dis}}\mathcal{L}_{\text{dis}},
 \end{aligned}
  \label{eq:3.11}
\end{equation}
\noindent where $\lambda_{\text{dis}}$ is a hyperparameter that controls the importance of the knowledge distillation loss $\mathcal{L}_{\text{dis}}$ during training.

\section{Experiments}
\label{sec:exp}

\subsection{Experimental setup and details}

\textbf{Experimental setup.} Following \cite{ze2023gnfactor, lu2024manigaussian}, we select 10 representative manipulation skills in RLBench\cite{james2020rlbench} and 5 daily manipulation skills in the real world for our experiments. Each skill has at least two variations and 20 demonstrations during training that cover multiple types, such as position, shape, and color. To achieve a high success rate for these skills, the manipulation policy needs to learn generalizable knowledge rather than overfitting the limited given demonstrations. For visual observation, we only use the front RGB-D image with $128\times128$ resolutions. 
To demonstrate the performance of our method under different incremental settings, we define several configurations, represented as \textcolor{blue}{\textbf{Bn-kNm}}. This notation indicates that the policy is initially trained on \textcolor{blue}{\textbf{n}} base skills, followed by the addition of \textcolor{blue}{\textbf{m}} new skills in each step, with a total of \textcolor{blue}{\textbf{k}} steps.
\\
\textbf{Evaluation Metric.} We report the performance of the agent on each learned skill by the average success rate. At each incremental step, we present the average success rate for old, new, and all (combined old and new) skills. In the simulation, we evaluate the agent with 25 episodes per skill, whereas in the real world, we use 10 episodes per skill. During evaluation, the agent continues to take actions until an oracle signals task completion or the agent reaches a maximum of 25 steps.\\
\textbf{Implementation Details.} For model design, we use different encoders to transform corresponding modality data into tokens, which serve as the input for the Extendable PerceiverIO. 
\begin{table}[t]
\centering
\begin{tabular}{c|ccc|cc}
\Xhline{1pt}
&TRS & EPIO & DIS & B5-1N1 & B5-5N1 \\
\hline
      R1     &      &       &            & 20.7 & 5.2\\
R2 &  \checkmark   &       &        & 49.3 & 27.6 \\
R3 & \checkmark &  \checkmark     &     & 54.0 & 32.4 \\
Ours & \checkmark & \checkmark &\checkmark & 56.7 & 36.0  \\
\Xhline{1pt}
\end{tabular}
\caption{Ablation Study on two experiment setup. We report the average success rate of all learned skills.}
\vspace{-3mm}
\label{tab:ablation study}
\end{table}
The RGB-D images are projected and transformed into voxels, which are then encoded by a 3D convolutional encoder with a UNet architecture, while text instructions are encoded using CLIP RN50~\cite{radford2021learning}, and the proprioception data is encoded by a single-layer MLP. After encoding, the tokens from all modalities have the same dimension of 512. The hyperparameter $\lambda_{\text{dis}}$ is set as 0.01 and the action prompt length is 16. We store 2 keyframe replays of the total 20 demonstrations of learned skills and train the agent on two NVIDIA RTX 4090 GPUs with a batch size of 1, a learning rate of 0.002, and 100k iterations. 
More studies about the hyperparameters are shown in the Appendix.

\subsection{Simulation results}
\label{sec4.2}
\begin{table*}[h]
\centering
\begin{tabular}{c|c|c|cccccccccc}
\Xhline{1pt}
\multicolumn{1}{c|}{\multirow{2}{*}{Frozen layer}} & \multicolumn{1}{c|}{\multirow{1}{*}{Convergence }} &
  \multicolumn{1}{c|}{Trained } &
  
  \multicolumn{2}{c}{Slide block} &
  \multicolumn{2}{c}{Put in drawer} &
  \multicolumn{2}{c}{Drag stick} &
  \multicolumn{2}{c}{Push buttons} &
  \multicolumn{2}{c}{Stack blocks} \\ \cline{4-13} 
\multicolumn{1}{c|}{} &\multicolumn{1}{c|}{steps} &
  \multicolumn{1}{c|}{param} &
  
  Old &
  \multicolumn{1}{c}{New} &
  Old &
  \multicolumn{1}{c}{New} &
  Old &
  \multicolumn{1}{c}{New} &
  Old &
  \multicolumn{1}{c}{New} &
  Old &
  New \\ \hline
Non-frozen  & 100000 & 47M   & 43.2 & \textbf{60.0}    & 44.8   & \textbf{16.0}  & 40.8  & \textbf{92.0}  & 44.0  & \textbf{28.0}  & 45.6   & \textbf{12.0}  \\
Encoder   & 75000  & 37M   & 50.4  & 54.0  & 51.2 & 12.0    & 45.6   & 88.0  & 48.4  & 24.0  & 52..0   & 8.0  \\
EPIO & 70000 & 18M & 52.0  & 52.0 & 52.8  & 16.0   & 46.4     & 84.0  & 50.4   & 24.0   & 49.6  & 8.0    \\
Decoder   & 75000  & 39M   & 45.6  & 24.0    & 47.2   & 0.0  & 42.4   & 40.0    & 44.8   & 4.0   & 46.4  & 0.0   \\
Encoder+EPIO & 60000   & 8M   & \textbf{57.6} & 52.0   & \textbf{56.8}   & 12.0 & \textbf{50.4}   & 84.0  & \textbf{55.2}  & 20.0 & \textbf{56.0}  & 8.0 \\ \Xhline{1pt}
\end{tabular}
\caption{
Performance of five sets B5-1N1 experiments with the same base skills and different new skills, while freezing different network layers. For each new skill, we train a total of 100k iterations and report the average success rate and the average model convergence steps.
}
\label{tab:frozen}
\end{table*}
\textbf{Performance comparison with different methods.} We conduct the skill-incremental learning experiment in the B5-5N1 setting, where we first train the policy on five base skills and then gradually learn a new skill at each subsequent step with a total of five steps. 
\begin{table}[ht]
\centering
\begin{tabular}{ccccc}
\Xhline{1pt}
Methods & B5-1N5 & B2-4N2 & B3-2N3  \\
\hline
ManiGaussian~\cite{lu2024manigaussian} & 25.6 & 10.4 & 17.3 \\
M-TIB~\cite{lu2024manigaussian,rebuffi2017icarl,castro2018end}        & 30.8 & 28.4 & 33.3 \\
Ours (iManip)       & \textbf{37.2} & \textbf{36.8} & \textbf{41.3} \\
\Xhline{1pt}
\end{tabular}
\caption{Average success rate of all learned skills on different skill-incremental setup.}\vspace{-3mm}
\label{tab:task-incremental setup}
\end{table}
To demonstrate the performance of our method in robotic skill-incremental learning, we compare with two standard multi-task manipulation policies, PecAct~\cite{shridhar2023perceiver} and ManiGaussian~\cite{lu2024manigaussian},
by retraining the agent to learn new manipulation skills.
Furthermore, we apply two Traditional Incremental Baselines (TIB) for visual classification to the above policies for comparison, termed P-TIB and M-TIB respectively.
As shown in \Cref{tab:maintable}, the results demonstrate that the performance of our method significantly outperforms the others at each subsequent step. This demonstrates that our method better facilitates the learning of new skills while mitigating the forgetting of previous skills.
More detailed results of each learning skill at every step are shown in the Appendix.
\\
\textbf{Ablation Study.} We conduct the ablation study to validate the effectiveness of each policy, as shown in \Cref{tab:ablation study}. R1 is the control group where the agent does not have any incremental policy. When we add the Temporal Replay Strategy (\textbf{TRS}) to the agent, the setup of B5-1N1 and B5-5N1 improve by 28.6\% and 22.4\% in the success rate, respectively (see R2). The significant performance improvement stems from the success of our temporal replay strategy that maintains the integrity of the temporal data. 

When we add Extendable PerceiverIO (\textbf{EPIO}) to the agent, the success rates of B5-1N1 and B5-5N1 further improve by 4.7\% and 4.8\%, respectively (see R3). The EPIO design works because the skill-specific action prompts help the agents incrementally learn action primitives for new skills, and the extendable weights designed in transformer blocks allow the model to preserve the old knowledge while adapting to new skills. Our complete policy achieves the best success rate, where the Distillation mechanism (\textbf{DIS}) improves the performance by 2.7\% and 3.6\%, respectively. 

Through the ablation study, we find that the replay policy has the greatest impact on overall performance. Without old data for retraining, the agent is more likely to forget previously learned knowledge. This occurs because data plays a crucial role in robotic manipulation, and without the support of previous data, the agent is prone to overfitting the data of new skills.
\\
\textbf{Effect of parameter freezing on skill-incremental learning.} 
The agent consists of three main components: the encoders, the extendable PerceiverIO, and the policy decoder. 
we freeze each component individually to evaluate its effect.
As shown in \Cref{tab:frozen},
five sets B5-1N1 experiments on different new skills demonstrate the following: 
(1) Freezing the encoders or the extendable PerceiverIO helps retain knowledge from the old skills without significantly hindering the learning of the new skill (Lines 1,2,3).
(2) The policy decoder is crucial for learning new skills (Lines 1,4). 
(3) Freezing the above components helps decrease the number of parameters and accelerate convergence.

Based on these findings, we freeze both the encoder and the extendable PerceiverIO, leaving only the decoder with newly appended action prompts and weights for new skill training, achieving faster convergence, fewer parameters, and better performance (Line 5)!
\begin{table*}[ht]
\centering
\begin{tabular}{c|cccccccccc}
\Xhline{1pt}
\multirow{2}{*}{Manipulation skills} & \multicolumn{2}{c}{Base} & \multicolumn{2}{c}{Step 1} & \multicolumn{2}{c}{Step 2} & \multicolumn{2}{c}{Step 3} & \multicolumn{2}{c}{Step 4} \\ \cline{2-11}
                                     & BL      & Ours     & BL       & Ours      & BL      & Ours       & BL      & Ours       & BL       & Ours      \\ \hline
Slide toy                  & 90.0          & 90.0     & 10.0           & 80.0      & 0             & 80.0       & 0             & 60.0       & 0              & 60.0      \\
Open drawer                          & -             & -        &\cellcolor{yellow!20} 70.0  &\cellcolor{yellow!20} 60.0      & 0             & 60.0       & 0             & 50.0       & 0              & 40.0      \\
Pick and place                       & -             & -        & -              & -         & \cellcolor{yellow!20} 60.0       & \cellcolor{yellow!20} 60.0       & 0             & 60.0       & 0              & 50.0      \\
Pour water                           & -             & -        & -              & -         & -             & -          & \cellcolor{yellow!20} 40.0          & \cellcolor{yellow!20} 40.0       & 0              & 10.0      \\
Close jar                            & -             & -        & -              & -         & -             & -          & -             & -          & \cellcolor{yellow!20} 50.0           &  \cellcolor{yellow!20}40.0      \\ \hline
\rowcolor{gray!10} 
Old manipulation skills                                 & 90.0          & 90.0     & 10.0           & \textcolor{red}{\textbf{80.0}} \textcolor{red}{\scalebox{0.8}{+70.0}}      & 0       & \textcolor{red}{\textbf{70.0}}  \textcolor{red}{\scalebox{0.8}{+70.0}}      & 0        & \textcolor{red}{\textbf{56.7}}  \textcolor{red}{\scalebox{0.8}{+56.7}}       & 0             & \textcolor{red}{\textbf{40.0}}  \textcolor{red}{\scalebox{0.8}{+40.0}}     \\
\rowcolor{gray!10} 
All manipulation skills                                 & 90.0          & 90.0     & 40.0           & \textcolor{red}{\textbf{70.0}}  \textcolor{red}{\scalebox{0.8}{+30.0}}      & 20.0    & \textcolor{red}{\textbf{66.7}}  \textcolor{red}{\scalebox{0.8}{+46.7}}      & 10.0     & \textcolor{red}{\textbf{52.5}}  \textcolor{red}{\scalebox{0.8}{+42.5}}   & 10.0     & \textcolor{red}{\textbf{40.0}}  \textcolor{red}{\scalebox{0.8}{+30.0}}       \\ \Xhline{1pt}
\end{tabular}
\caption{Real world experiments. The table reports the success rate of BaseLine (BL) and Ours. \textcolor{red}{\scalebox{0.8}{+num}} is the improvement of our method compared to the baseline.}
\label{tab:real world exp}
\end{table*}
\begin{figure}[t]
  \centering
   \includegraphics[width=0.96\columnwidth]{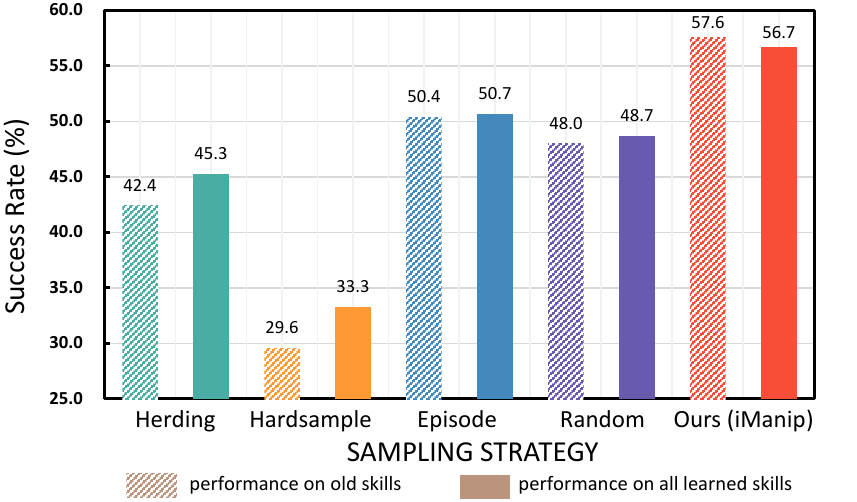}
   \caption{Average success rate on different replay methods.
   }\vspace{-5mm}
   \label{fig:replaymethod}
\end{figure}
\\
\textbf{Experiments on different skill-incremental setup.} We implement different skill-incremental settings to validate the generalizability of our method and report the average success rate across all previously learned skills after the last incremental step. As shown in \Cref{tab:task-incremental setup}, compared to ManiGaussian and M-TIB, our method achieves higher success rates in all incremental experimental settings. This shows that our approach has stronger performance and better generalization capabilities.
\\
\textbf{Exploring data replay methods.} We compare our temporal replay strategy with the classical rehearsal-based method on the setup of B5-1N1, as shown in \Cref{fig:replaymethod}. \textbf{Herding}~\cite{rebuffi2017icarl} and \textbf{Hardsample}~\cite{chaudhry2018riemannian} are two methods for selecting the most representative samples from the data. \textbf{Episode} refers to replaying a complete trajectory. \textbf{Random} refers to random sampling. The results show that classic herding sampling and hard-exemplar sampling perform poorly on old skills due to neglecting temporal integrity in robotic demonstrations. In contrast, replaying complete trajectories or random sampling better preserves the temporal integrity of the samples, leading to better performance. Our temporal replay strategy, leveraging the farthest-distance entropy sampling for each keyframe can sample more different variants and achieves the best success rate.
\begin{figure}[t]
  \centering
   \includegraphics[width=0.96\columnwidth]{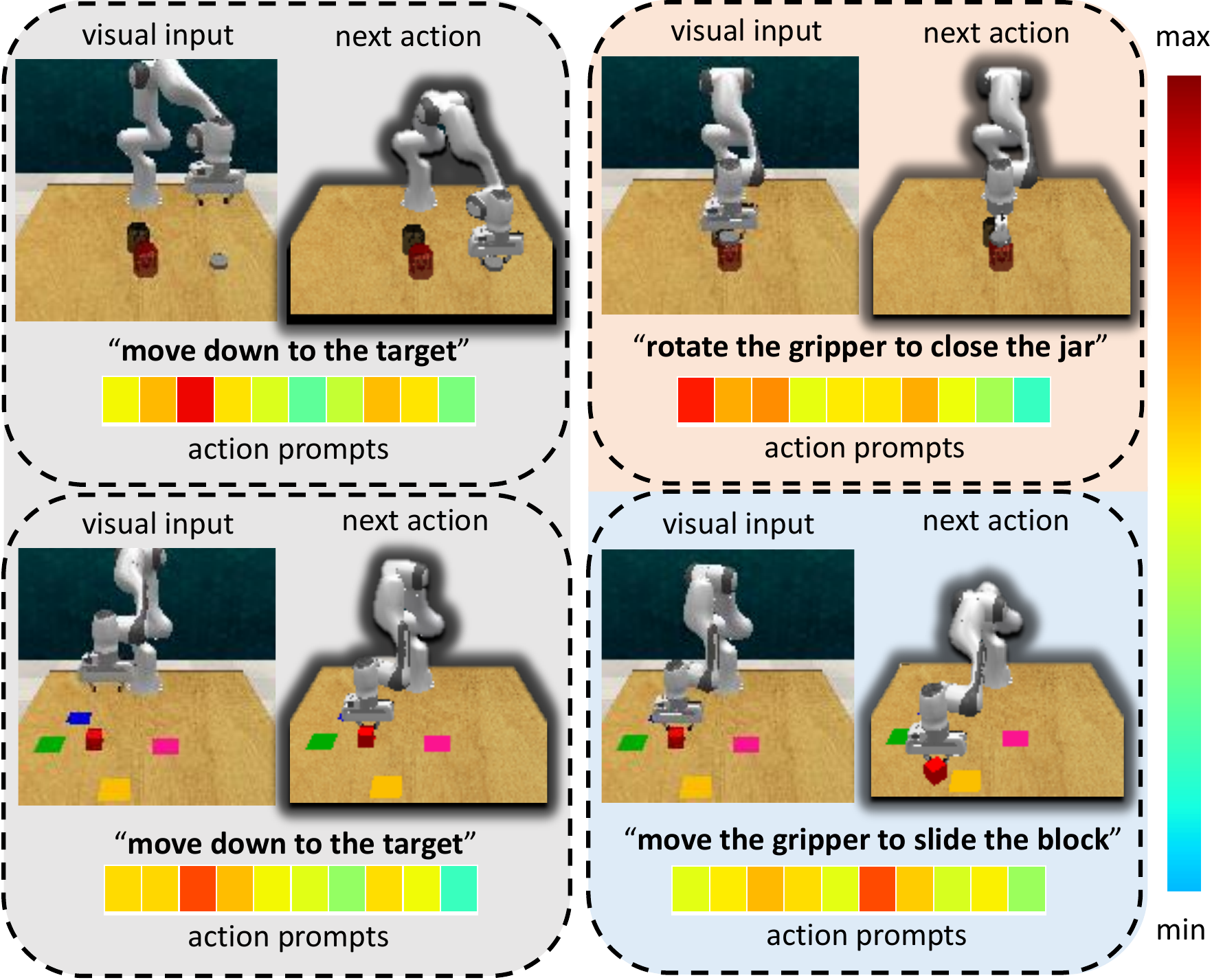}
   \caption{Visualization of skill-specific action prompts by Grad-CAM as the agent executes two manipulation skills: close jar (the first row) and slide block (the second row). 
   }
   \label{fig:visual} \vspace{-6mm}
\end{figure}
\\
\textbf{Visualization of the skill-specific action prompts.} 
We visualize our skill-specific action prompts by Grad-CAM~\cite{selvaraju2017grad}, as shown in ~\Cref{fig:visual}. Our experiments train the agent in the B5-5N1 setting with totaling 10 action prompts.
We first compute the sum of the parameter gradients for each action prompt and then normalize these values to calculate Grad-CAM weights, displayed in different colors.  
We show the results across two different skills.
When the agent executes the same action, \eg ``move down to the target'', the third action prompt weight is maximized (see the first column). 
Furthermore, when performing different actions, different weights of skill-specific action prompts are maximized (see the second column). It demonstrates that action prompts can learn skill-specific action primitives. Freezing old action prompts while learning new ones helps prevent forgetting and adapt to new action primitives.

\subsection{Real world experiments}
We conduct five manipulation skills in the real-world environment to further validate the effectiveness of our method. We use a Franka Panda robotic arm to execute the action and a Realsense D455 camera to capture RGB-D images as observations. For each training step, we collect 20 demonstrations. The training setup is B1-4N1 where we first train on a base skill, then incrementally add one new skill at a time for training, with a total of four new skills added. During testing, we perform 10 test runs for each learned skill and report the success rate of task execution. \textbf{More details about the real world experiments and videos are shown in the supplementary material.}

We compare our method with the baseline \cite{lu2024manigaussian} without any incremental policy. As shown in \Cref{tab:real world exp}, it is evident that without the incremental strategy, the knowledge of previous skills is rapidly forgotten when training on new skills. 
After incorporating our incremental strategy, the success rate on new skills is lower than the baseline. This occurs because the baseline has overfitted to the new skill, while our model, which is designed to retain knowledge from previous skills, experiences a slight decrease in its ability to learn new skills. This trade-off is an inherent challenge in incremental learning.

\section{Conclusion}
In this work, we focus on a new and challenging setting, skill-incremental learning in robotic manipulation, which is to continually learn new skills while maintaining the previously learned skills. We conduct experiments on the RLBench benchmark and find that traditional methods suffer from catastrophic forgetting because they overlook the temporal and action complexities of robotic manipulation. Our approach proposes a temporal replay strategy to address the temporal complexities and an extendable PerceiverIO model with adaptive action prompts to address the action complexities. 
Extensive experiments demonstrate that our iManip framework excels in effectiveness, robustness, lightweight design, and extendability.



\clearpage
\setcounter{page}{1}
\maketitlesupplementary

\begin{figure*}[h]
  \centering
   \includegraphics[width=2\columnwidth]{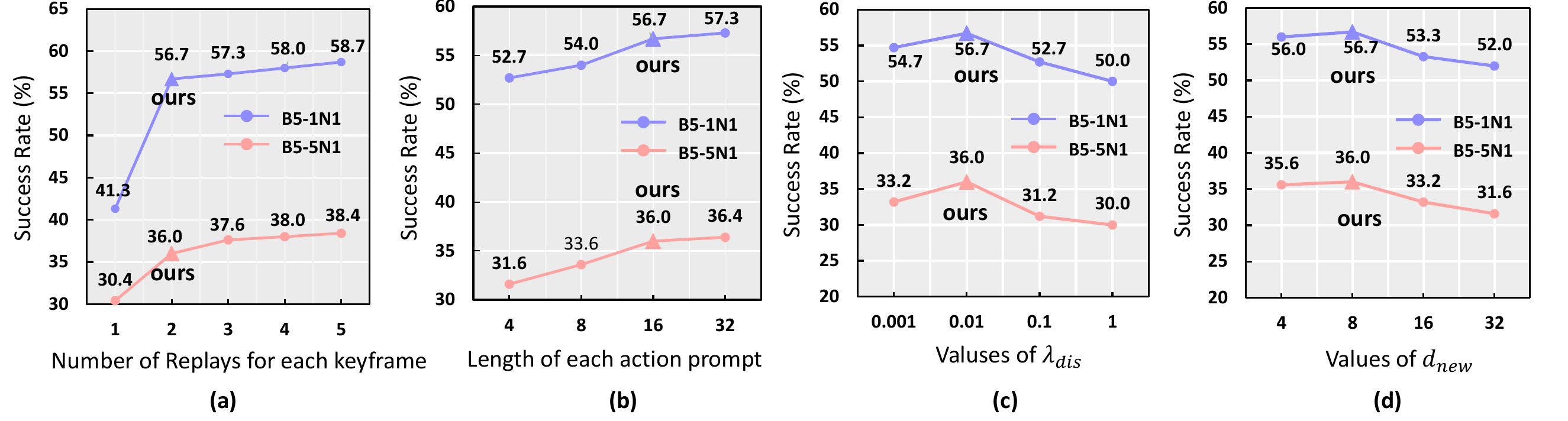}
   \caption{Experiments results of four key hyperparameters.
   }\vspace{-2mm}
   \label{fig:hys}
\end{figure*}

\section{Experiments on key hyperparameters}
\label{sec:rationale}
We conducte experiments in RLbench on four key hyperparameter: the number of replays per keyframe, the length of each action prompt, the hyperparameter of distillation loss factor $\lambda_{dis}$, and the dimension of the appended PerceiverIO weight matrix $d_{new}$, using the B5-1N1 and B5-5N1 setups. \\
\textbf{Exploring the amount of replays per keyframe.} 
For each new manipulation skill, 20 training trajectories are used for training. Specifically, as shown in ~\Cref{fig:hys} (a), we vary the replay size from 1 to 5 per keyframe based on the temporal replay strategy. The results demonstrate that increasing the number of replays leads to improved model performance. For memory efficiency, We store 2 samples per old keyframe for replay. \\
\textbf{Exploring the length of each action prompt.} 
During the training in the new incremental step, there are new skill-specific action prompts assigned to learn action primitives. 
Specifically, as shown in ~\Cref{fig:hys} (b), we conduct experiments with action prompt lengths varying from 4 to 32. The results show that extending the length of skill-specific action prompts enhances model performance. For better memory efficiency, the action prompt length is set to 16. \\
\textbf{Exploring the distillation loss factor $\lambda_{dis}$.} 
In our iManip, we propose to use a distillation loss with a factor $\lambda_{dis}$ to transfer the knowledge from the old model to the agent. 
We conduct ablation experiments with different values of $\lambda_{dis}$ to investigate the impact of the distillation loss on model performance, as shown in ~\Cref{fig:hys} (c). Results indicate that distillation from the old model contributes to improving agent performance. However, a larger weight of $\lambda_{dis}$ causes the model to focus too much on old skills, negatively affecting overall performance. 
We set this parameter to 0.001 to achieve optimal performance.\\
\textbf{Exploring the impact of expended PerceiverIO.} 
We propose extending the weights of the PerceiverIO to adapt to learn new action primitives. 
As shown in ~\Cref{fig:hys} (d), we perform ablation experiments on the dimension of the appended PerceiverIO weight matrix $d_{new}$. 
The results indicate that the best performance is achieved when $d_{new}$ is set to 8, and larger values hinder overall performance. This is because more new parameters for new skills training can interfere with the retention of old knowledge, leading to forgetting.

\section{More exploratory experiments}

\textbf{The effect of skill learning order.} 
Long-horizon skills are more challenging for robotic manipulation~\cite{garcia2024towards}.
Based on the number of keyframes, we organized robotic skills into three progressively more challenging levels, \ie short, medium, and long.
Skills with fewer than five keyframes are considered short-horizon, those with 5 to 10 keyframes are classified as medium, and skills with more than 10 keyframes are regarded as long-horizon. 
We conducted experiments on the learning sequence of skills with varying levels in the B2-2N2 setting, as shown in \Cref{tab:task order}. 
It is evident that, regardless of the skill learning sequence, our method maintains a balanced average accuracy after completing all skills in the final stage. This highlights the robustness of our approach, which can effectively adapt to learn different new skills. Furthermore, the results reveal that, compared to short-horizon skills, longer-horizon skills are more effective in acquiring general knowledge, thereby mitigating forgetting.
\begin{table}[t]
\centering
\begin{tabular}{c|c|cc|cccc}
\Xhline{1pt}
\multirow{2}{*}{Order} & \multirow{2}{*}{Base} & \multicolumn{2}{c|}{Step 1} & \multicolumn{4}{c}{Step 2} \\
      &    & B  & S1 &B  & S1 & S2 &Average \\ \hline
S-M-L & 72 & 42 & 44 & 42 & 36 & 10 & \textcolor{red}{29.3}\\
S-L-M & 72 & 44 & 10 & 40 & 8 & 44 & \textcolor{red}{30.7} \\
M-S-L & 48 & 42 & 66 & 34 & 42 & 10 & \textcolor{red}{28.7}\\
M-L-S & 48 & 38 & 8  & 32 & 4  & 58 & \textcolor{red}{31.3} \\
L-S-M & 10 & 6  & 60 & 4  & 46 & 38 &\textcolor{red}{29.3} \\
L-M-S & 10 & 8  & 48 & 4  & 34 & 52 &\textcolor{red}{30.0} \\ \Xhline{1pt}
\end{tabular}
\caption{Average success rate of skills with varying levels (Short, Medium, and Long) on different continuous learning orders. The new skills learned in the base step, 1st step, and 2nd step are termed B,S1, and S2 respectively.}
\label{tab:task order}
\vspace{-2mm}
\end{table}
\begin{table}[h]
\centering
\begin{tabular}{cc|cc}
\Xhline{1pt}
\multicolumn{2}{c|}{Method}                            & B5-1N5 & B3-2N3 \\ \hline
\multicolumn{1}{c|}{\multirow{3}{*}{Peract~\cite{shridhar2023perceiver}}}   & None  & 15.6   & 8.4    \\
\multicolumn{1}{c|}{}                          & +TIB  & 20.4   & 22.6   \\
\multicolumn{1}{c|}{}                          & +Ours & 25.6   & 30.7   \\ \hline
\multicolumn{1}{c|}{\multirow{3}{*}{GNFactor~\cite{ze2023gnfactor}}} & None  & 20.4   & 15.6   \\
\multicolumn{1}{c|}{}                          & +TIB  & 27.4   & 29.3   \\
\multicolumn{1}{c|}{}                          & +Ours & 33.6   & 36.9   \\ \hline
\multicolumn{1}{c|}{\multirow{3}{*}{3DDA~\cite{ke20243d}}}     & None  & 42.4   & 31.6   \\
\multicolumn{1}{c|}{}                          & +TIB  & 67.6   & 68.4   \\
\multicolumn{1}{c|}{}                          & +Ours & 72.8   & 76.4     \\ \Xhline{1pt}
\end{tabular}
\caption{The results of adapting our iManip to different robotic manipulation pipelines.}
\label{tab:plug and play}
\end{table} 
\begin{table*}[h]
\centering
\begin{tabular}{ccccc}
\Xhline{1pt}
\textbf{Manipulation skill}           & \textbf{Type}   & \textbf{Variations} & \textbf{Keyframes} & \textbf{Instruction Template}         \\ \hline
close jar               & color           & 20                  & 6.0                 & ``close the \_jar''                   \\ 
open drawer             & placement       & 3                   & 3.0                 & ``open the \_drawer''                 \\ 
sweep to dustpan        & size            & 2                   & 4.6                 & ``sweep dirt to the \_dustpan''       \\ 
meat off grill          & category        & 2                   & 5.0                 & ``take the \_off the grill''         \\ 
turn tap                & placement       & 2                   & 2.0                 & ``turn \_tap''                       \\ 
slide block             & color           & 4                   & 4.7                 & ``slide the block to \_target''      \\ 
put in drawer           & placement       & 3                   & 12.0                & ``put the item in the \_drawer''     \\ 
drag stick              & color           & 20                  & 6.0                 & ``use the stick to drag the cube onto the \_target'' \\ 
push buttons            & color           & 50                  & 3.8                 & ``push the \_button, [then the \_button]'' \\ 
stack blocks            & color, count    & 60                  & 14.6                & ``stack \_blocks''                   \\ \Xhline{1pt}
\end{tabular}
\caption{Task Information Table}
\label{tab:rlbench}
\end{table*} 
\begin{table*}[h]
\centering
\begin{tabular}{ccccccc}
\Xhline{1pt}
Robotic skills   & Base & Step 1 & Step 2 & Step 3 & Step 4 & Step 5 \\ \hline
close jar        & 28   & 24     & 40     & 32     & 16     & 20     \\
open drawer      & 56   & 72     & 64     & 68     & 60     & 64     \\
sweep to dustpan & 52   & 52     & 56     & 32     & 40     & 36     \\
meat off grill   & 80   & 76     & 52     & 60     & 52     & 52     \\
turn tap         & 64   & 64     & 60     & 60     & 56     & 56     \\
slide block      & -    & \textcolor{red}{52}     & 32     & 32     & 44     & 40     \\
put in drawer    & -    & -      & \textcolor{red}{32}     & 32     & 4      & 8      \\
drag stick       & -    & -      & -      & \textcolor{red}{64}     & 64    & 60     \\
push buttons     & -    & -      & -      & -      & \textcolor{red}{16}     & 12     \\
stack blocks     & -    & -      & -      & -      & -      & \textcolor{red}{12}     \\ \hline
Old              & 56.0 & 57.6   & 50.7   & 45.1   & 42.0   & 38.7   \\
All              & 56.0 & 56.7   & 48.0   & 47.5   & 39.1   & 36.0   \\ \Xhline{1pt}
\end{tabular}
\caption{Performance of iManip in the setup of B5-5N1.}
\label{tab:all}
\end{table*}
\\
\textbf{Plug and play.} Our skill-incremental policy can also be seamlessly integrated into other robotic multi-task learning pipelines. We apply our method to three robotic multi-task learning frameworks including Peract\cite{shridhar2023perceiver}, GNFactor\cite{ze2023gnfactor}, and 3DDA\cite{ke20243d}. 
we compare our method with pipelines that either did not include incremental methods or used traditional incremental learning methods. 
Specifically, for storing old replay data, all the aforementioned pipelines can utilize our temporal replay strategy to sample informative, temporally balanced samples from previous manipulation skills.
Furthermore, we can modify the transformer with our extensible self-attention layer by appending action prompts and incorporating a minimal number of trainable parameters to adapt to different action primitives. Notably, following the original setup of each paper, we use 20 demonstrations per manipulation skill for PerAct and GNFactor, and 100 demonstrations for 3DDA.
As shown in \Cref{tab:plug and play}, in each pipeline, our method achieves the highest task success rates. This demonstrates the excellent scalability of our incremental strategy.
\label{sec:rationale}
\begin{figure*}[t]
  \centering
   \includegraphics[width=2\columnwidth]{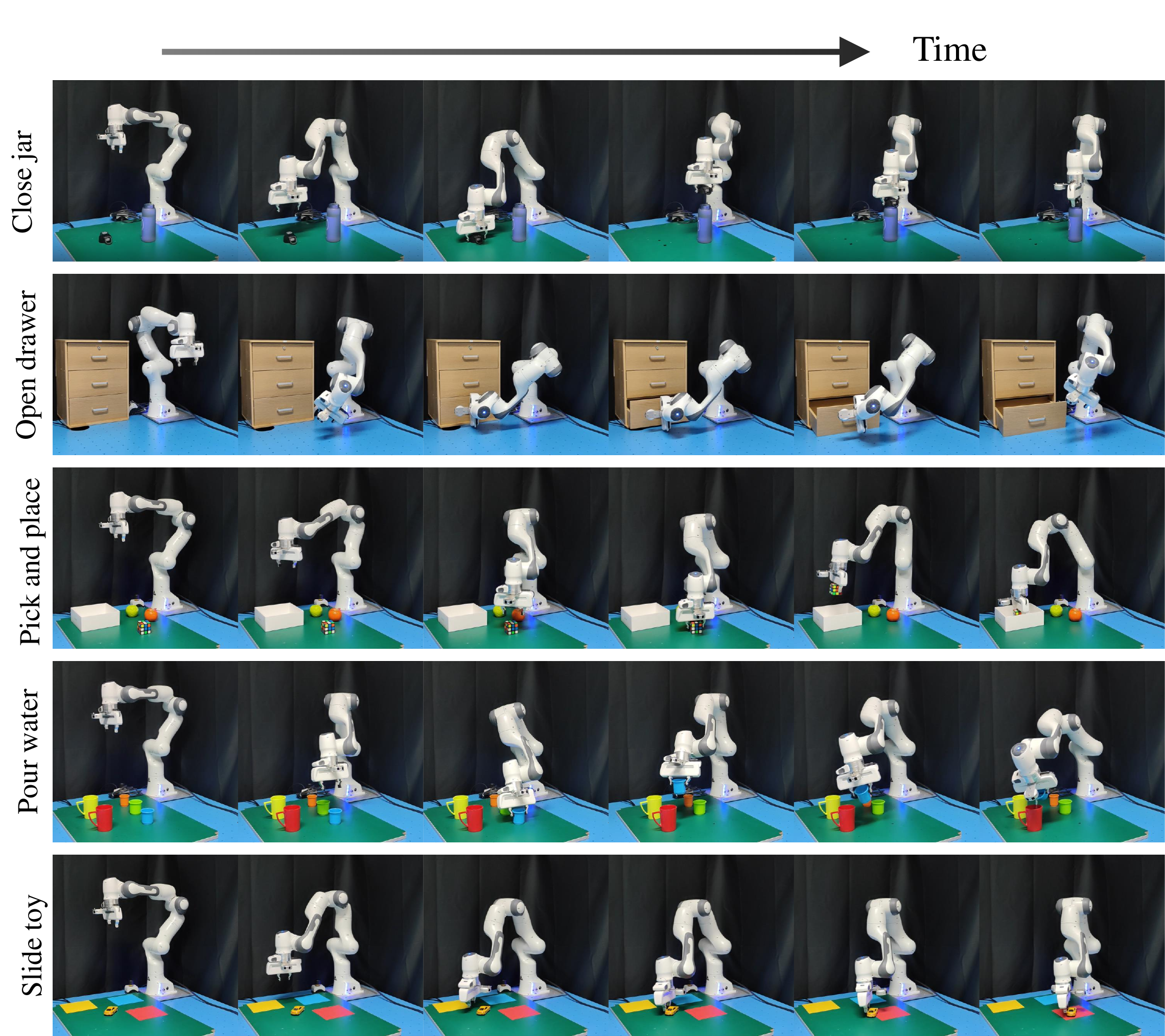}
   \caption{Keyframes for real robot manipulation skills.
   }
   \label{fig:realexp}
\end{figure*}

\section{More details about simulation experiments}
\textbf{Manipulation skills in rlbench.} We select 10 language-conditioned skills from RLBench \cite{james2020rlbench}, each involving at least two variations. An overview of these skills can be found in Table \ref{tab:rlbench}. The variations include random sampling of object colors, sizes, quantities, placements, and categories, resulting in a total of 166 distinct combinations. The color set consists of 20 different colors: red, maroon, lime, green, blue, navy, yellow, cyan, magenta, silver, gray, orange, olive, purple, teal, azure, violet, rose, black, and white. The size set includes two options: short and tall. The count set has three possible values: 1, 2, or 3. The placements and object categories are skill-specific. For instance, the "open drawer" skill has three placement options: top, middle, and bottom. Additionally, objects are randomly placed on the tabletop in various poses within a defined range.\\
\textbf{The experimental details of iManip in the B5-5N1 setup.} In this setup, the agent is first trained on the 5 base skills, and then a new skill is added at each step, with a total of 5 steps. The base skill includes \textit{close jar}, \textit{open drawer}, \textit{sweep to dustpan}, \textit{meat off grill} and \textit{turn tap}. The training sequence for the new skills is \textit{slide block}, \textit{put in drawer}, \textit{drag stick}, \textit{push buttons}, \textit{stack blocks}. Table \ref{tab:all} shows the success rate of each skill at each step. The Old is the average success rate of the old skills. For example, the Old in step 1 is the average success rate of the five base skills. The All is the average success rate of all learned skills at that step.

\section{More details about real world experiments}
In the real world experiment, we use the B1-4N1 setup, which allows the agent to gradually learn five different skills one by one. The five manipulation skills includes \textit{Silde toy to target}, \textit{Open drawer}, \textit{Pick and place}, \textit{Pour water} and \textit{Close jar}. Concretely, 
The skill of \textit{silde toy to target} requires the agent to move the toy to the color area specified by the instruction. The skill of \textit{Open drawer} requires the agent to open the drawer at the corresponding position, including variations for the top, middle, and bottom positions. The skill of \textit{Pick and place} requires the agent to grasp the specified object and place it at the designated location. The skill of \textit{Pour water} requires the agent to pick up the water-filled cup of a specified color and pour the water into the mug of another specified color. Lastly, the skill of \textit{Close jar} requires the agent to grasp the bottle cap and screw it onto the bottle. We present the keyframes of the five manipulation skills in sequence in Figure \ref{fig:realexp}, visually illustrating the specific steps of the execution process.

\section{Model architecture}
\textbf{Voxel Encoder}: We employ a compact 3D UNet with only 0.3M parameters to encode the input voxel of size $100^3\times 10$ (which includes RGB features, coordinates, indices, and occupancy) into our deep 3D volumetric representation, resulting in a size of $100^3\times128$. \\
\textbf{Original PerceiverIO.} The Extendable PerceiverIO in iManip is an improvement upon the original PerceiverIO \cite{jaegle2021perceiver}. A detailed explanation of the original PerceiverIO is provided here to offer a more comprehensive understanding of our Extendable PerceiverIO. 

The original PerceiverIO consists of 6 attention blocks designed to process sequences from multiple modalities (such as 3D volumes, language tokens, and robot proprioception) and output a corresponding sequence. To efficiently handle long sequences, the Perceiver Transformer uses a small set of latents to attend to the input, improving computational efficiency. The resulting output sequence is then reshaped back into a voxel representation to predict the robot's actions. The Q-function for translation is predicted using a 3D convolutional layer. For predicting openness, collision avoidance, and rotation, we apply global max pooling and spatial softmax to aggregate 3D volume features, and then project the aggregated feature to the output dimension using a multi-layer perceptron. \\
\textbf{Model inference.} The agent obtains the current observation state, including RGB-D, proprioception, and textual instructions, and predicts the end-effector pose for the next keyframe. It then uses a predefined motion planner (e.g., RRT-Connect) to solve for the motion path and joint control angles. This approach reduces the sequential decision-making problem to predicting the optimal keyframe action for the next step based on the current observation.

\end{document}